%% file: ADMM_GL_inf_4.tex
\documentclass[10pt,letterpaper,twoside]{article}

\usepackage{amsmath,amssymb,amsthm}
\usepackage{url}
\usepackage{fullpage}
\usepackage{natbib}
\usepackage[boxed]{algorithm2e}
\usepackage{color}
\usepackage{hyperref}

\hypersetup{
colorlinks,
citecolor=magenta,
linkcolor=blue,
urlcolor=blue}

\usepackage{subfigure}
\usepackage{graphicx}
\input{./macros/macro_general}


\newcommand{\Prox}{\mbox{Prox}}
\newcommand{\Thetahat}{\hat{\var}}
\newcommand{\Gammahat}{\hat{\Gamma}}

\title{ADMM Algorithm for Graphical Lasso\\ with an $\ell_\infty$ Element-wise Norm Constraint}
\author{Karthik Mohan}
\date{\today}

\begin{document}

\maketitle
\begin{abstract}
We consider the problem of Graphical lasso with an additional $\ell_{\infty}$ element-wise norm constraint on the precision matrix. This problem has applications in high-dimensional covariance decomposition such as in \citep{Janzamin-12}. We propose an ADMM algorithm to solve this problem. We also use a continuation strategy on the penalty parameter to have a fast implemenation of the algorithm.
\end{abstract}

\section{Problem}
The graphical lasso formulation with $\ell_{\infty}$ element-wise norm constraint is as follows:
\begin{eqnarray} \label{eq:GL_inf}
\begin{array}{rc}
\MIN_{\var \in \R^{p \times p}, \var \succ 0} & -\log\det(\var) + \langle \tS, \var \rangle + \gamma \|\var - \diag(\var)\|_1 \\
\mbox{s.t.} & \|\var - \diag(\var)\|_{\infty} \leq \lambda,
\end{array}
\end{eqnarray}
where $\|\cdot\|_1$ denotes the $\ell_1$ norm, and $\|\cdot\|_\infty$ denotes the $\ell_\infty$ element-wise norm of a matrix. For a matrix $X$, $\|X\|_{\infty} = \displaystyle \max_{i,j}|X_{ij}|$.
This formulation first appeared in \citep{Janzamin-12} in the context of high-dimensional covariance decomposition.
We next provide an efficient ADMM algorithm to solve (\ref{eq:GL_inf}).

\section{ADMM approach}
The \emph{alternating direction method of multipliers} (ADMM) algorithm \citep{Boyd-11,Eckstein-12} is especially suited to solve optimization problems whose objective can be decomposed into the sum of many \emph{simple} convex functions. By simple, we mean a function whose proximal operator can be computed efficiently. The proximal operator of a function $f$ is given by:
\begin{eqnarray} \label{eq:prox-f}
\Prox_f(\tA,\lambda) &=& \argmini_{X} \frac{1}{2}\|\tX - \tA\|_F^2 + \lambda f(\tX)
\end{eqnarray}

Consider the following optimization problem:

\begin{eqnarray} \label{eq:simple-problem}
\begin{array}{rc}
\MIN_{\tX, \tY } & f(\tX) + g(\tY) \\
\mbox{s.t.} & \tX = \tY,
\end{array}
\end{eqnarray}

where we assume that $f$ and $g$ are simple functions.

The ADMM algorithm alternatively optimizes the augmented Lagrangian to (\ref{eq:simple-problem}), which is given by:
\begin{eqnarray} \label{eq:aug_lag_GL_inf}
\Lcal_{\rho}(\tX,\tY,\Lambda) = f(\tX) + g(\tY) + \langle \Lambda, \tX - \tY \rangle + \frac{\rho}{2}\|\tX - \tY\|_F^2.
\end{eqnarray}

The $(k+1)th$ iteration of ADMM is then given by:

\begin{eqnarray} \label{eq:ADMM-updates}
\begin{array}{rcl}
\tX^{k+1} &\leftarrow& \argmini_{\tX} \Lcal_{\rho}(\tX, \tY^k,\Lambda^k) \\
\tY^{k+1} &\leftarrow& \argmini_{\tY} \Lcal_{\rho}(\tX^{k+1}, \tY, \Lambda^k) \\
\Lambda^{k+1} &\leftarrow& \Lambda^k + \rho(\tX^{k+1} - \tY^{k+1})
\end{array}
\end{eqnarray}

Note that each iteration in (\ref{eq:ADMM-updates}) has closed form updates if $f$ and $g$ have closed form proximal operators. The ADMM algorithm has a $O(1/\epsilon)$ convergence rate \citep{Goldfarb-10} just as for proximal gradient descent. We next reformulate our problem (\ref{eq:GL_inf}) and construct an ADMM algorithm.

\section{Reformulation by introducing new variables}
We now reformulate (\ref{eq:GL_inf}) into the standard form in (\ref{eq:simple-problem}) to derive an ADMM algorithm for our problem. We first define some notation.
Let,
\begin{eqnarray} \label{eq:X-Y-const}
\begin{array}{rcl}
\tX &=& \left[ \begin{array}{c} \var \\ \Gamma \end{array}\right] \in \mathbb{R}^{2p \times p} \\
\tY &=& \left[ \begin{array}{c} \Thetahat \\ \Gammahat \end{array}\right] \in \mathbb{R}^{2p \times p} \\
f(\tX) &=& -\log\det(\var) + \langle \tS, \var \rangle + \Ical_{\{\var \succ 0\}} + \gamma \|\Gamma - \diag(\Gamma) \|_1  \\
g(\tY) &=& \Ical_{\{\|\Thetahat - \diag(\Thetahat)\|_{\infty} \leq \lambda \}} + \Ical_{\{\Thetahat = \Gammahat\}},
\end{array}
\end{eqnarray}
where $\Ical_{\{.\}}$ denotes the indicator function that equals zero if the statement inside $\{.\}$ is true and $\infty$ otherwise.
Then note that (\ref{eq:GL_inf}) is equivalent to (\ref{eq:simple-problem}) with $\tX,\tY,f,g$ as in (\ref{eq:X-Y-const}).

\section{ADMM algorithm for Glasso with an $\ell_\infty$ element-wise norm constraint}
Define the following operators:
\begin{eqnarray}
\begin{array}{rcl}
\expand(\tA;\rho) &=& \argmini_{\var} -\log\det(\var) + \frac{\rho}{2}\|\var - \tA\|_F^2 \\
\Scal(\tA;\gamma) &=& \argmini_{\Gamma} \frac{1}{2}\|\Gamma - \tA\|_F^2 + \gamma \|\Gamma - \diag(\Gamma)\|_1 \\
\Pcal_{\infty}(\tA;\lambda) &=& \argmini_{\widetilde{\var}: \|\widetilde{\var} - \diag(\widetilde{\var})\|_{\infty} \leq \lambda} \|\widetilde{\var} - \tA\|_F^2
\end{array}
\end{eqnarray}
Plugging in the choice of $\tX, \tY$ and $f(\tX), g(\tY)$ from (\ref{eq:X-Y-const}) into (\ref{eq:ADMM-updates}), we get the following algorithm:

\begin{algorithm}[H] 
{\bf input}: $\rho > 0$ \;
{\bf Initialize}: Primal variables to the identity matrix and dual variables to the zero matrix\;
\While{Not converged}{ 
$\var \leftarrow \expand \left(\Thetahat - (\tS + \Lambda_{\var})/\rho; \rho \right)$ \;
$\Gamma \leftarrow \Scal \left(\Gammahat - \Lambda_{\Gamma}/\rho; \frac{\gamma}{\rho} \right)$ \;
$\Thetahat \leftarrow \Pcal_{\infty} \left( \frac{1}{2}(\var + \Gamma) + \frac{(\Lambda_{\var} + \Lambda_{\Gamma})}{2\rho} ; \lambda \right)$ \;
$\Gammahat = \Thetahat$ \;
$\Lambda_{\var} = \Lambda_{\var} + \rho(\var - \Thetahat)$ \;
$\Lambda_{\Gamma} = \Lambda_{\Gamma} + \rho(\Gamma - \Gammahat)$
} 
\caption{\label{algo:ADMM_GL_inf} \ADMM algorithm for graphical lasso with an additional $\ell_\infty$ norm constraint in (\ref{eq:GL_inf})}
\end{algorithm}
Note that we have 
\begin{align*}
\expand(\tA;\rho) &= \frac{\rho A +  (\rho^2 A^2 + 4 \rho I)^{1/2}}{2 \rho} \\
\Scal(\tA;\gamma) &= \text{Soft-Threshold}_\gamma (A) \\
\Pcal_{\infty}(\tA;\lambda) &= \text{Clip}_\lambda(A),
\end{align*}
where
\begin{align*}
\text{Soft-Threshold}_\gamma (A)_{ij} = 
\left\{ \begin{array}{ll}
A_{ij} & i=j \\
\sign(A_{ij})\max(|A_{ij}| - \gamma, 0) & i \neq j
\end{array} \right.
\end{align*}
and 
\begin{align*}
\text{Clip}_\lambda(A)_{ij} =
\left\{ \begin{array}{ll}
A_{ij} & i=j \\
\sign(A_{ij})\min(|A_{ij}|,\lambda) & i \neq j.
\end{array} \right.
\end{align*}

We add a continuation scheme to Algorithm \ref{algo:ADMM_GL_inf}, where $\rho$ is varied through the iterations. Specifically, we initially set $\rho = 1$ and double $\rho$ every 20 iterations. We terminate the algorithm when the relative error is small or when $\rho$ is too large as 
\begin{align*}
\frac{\|\Lambda_{\var}^{k+1} - \Lambda_{\var}^k\|_F}{\max(1,\|\Lambda_{\var}^k\|_F)} < \epsilon \quad \text{or}  \quad \rho > 10^6.
\end{align*}
We apply our algorithm to the problem of high-dimensional covariance decomposition, details of which can be found in \citep{Janzamin-arxiv-12}. 

\bibliographystyle{unsrtnat}
\bibliography{ADMM_GL_inf}
\end{document}

%% file: macros/macro_general.tex
\newcommand{\beq}{\begin{eqnarray}}
\newcommand{\eeq}{\end{eqnarray}}
\newcommand{\beqs}{\begin{eqnarray*}}
\newcommand{\eeqs}{\end{eqnarray*}}
\newcommand{\barr}{\begin{array}}
\newcommand{\earr}{\end{array}}
\newcommand{\beqa}{\begin{eqnarray} \begin{array}}
\newcommand{\eeqa}{\end{array} \end{eqnarray}}
\newcommand{\beqas}{\begin{eqnarray*} \begin{array}}
\newcommand{\eeqas}{\end{array} \end{eqnarray*}}

\newcommand{\bthm}{\begin{theo}}
\newcommand{\ethm}{\end{theo}}
\newcommand{\blem}{\begin{lem}}
\newcommand{\elem}{\end{lem}}
\newcommand{\bdefn}{\begin{defn}}
\newcommand{\edefn}{\end{defn}}

\newcommand{\MIN}{\displaystyle \min}

\newcommand{\sign}{\mathrm{sign}}

\newcommand{\tA}{\boldsymbol{A}}

\newcommand{\tS}{\boldsymbol{S}}

\newcommand{\tX}{\boldsymbol{X}}
\newcommand{\tY}{\boldsymbol{Y}}

\newcommand{\R}{\mathbb{R}}

\newcommand{\Ical}{\mathcal{I}}

\newcommand{\Lcal}{\mathcal{L}}

\newcommand{\Pcal}{\mathcal{P}}

\newcommand{\Scal}{\mathcal{S}}

\newcommand{\diag}{\mbox{diag}}

\newcommand{\var}{\boldsymbol{\Theta}}

\newcommand{\expand}{\mbox{Expand}}
\newcommand{\ADMM}{\mbox{ADMM }}

\newcommand{\argmini}{\displaystyle \mathop{\rm argmin}}

\definecolor{color1}{rgb}{0.11764705882,0.31372549019,1}
\definecolor{color2}{rgb}{0.25490196078,0.41176470588,1}
\definecolor{color3}{rgb}{0,0,1}
\definecolor{color4}{rgb}{0.11764705882,0.56470588235,1}
\definecolor{color5}{rgb}{1,0.49803921568,0.14117647058}
\definecolor{color6}{rgb}{1,0.07843137254,0.57647058823}
\definecolor{color7}{rgb}{1,0,0} 
\definecolor{color8}{rgb}{0.64705882352,0.16470588235,0.16470588235}
\definecolor{color9}{rgb}{0, 0.80392156862, 0}